# Simple black-box universal adversarial attacks on medical image classification based on deep neural networks

**Kazuki Koga[1], Kazuhiro Takemoto[1*]**
*1) Department of Bioscience and Bioinformatics, Kyushu Institute of Technology, Iizuka, Fukuoka 820-8502, Japan*
*\*Corresponding author's e-mail: takemoto@bio.kyutech.ac.jp*

## Abstract

Universal adversarial attacks, which hinder most deep neural network (DNN) tasks using only a small single perturbation called a universal adversarial perturbation (UAP), is a realistic security threat to the practical application of a DNN. In particular, such attacks cause serious problems in medical imaging. Given that computer-based systems are generally operated under a black-box condition in which only queries on inputs are allowed and outputs are accessible, the impact of UAPs seems to be limited because well-used algorithms for generating UAPs are limited to a white-box condition in which adversaries can access the model weights and loss gradients. Nevertheless, we demonstrate that UAPs are easily generatable using a relatively small dataset under black-box conditions. In particular, we propose a method for generating UAPs using a simple hill-climbing search based only on DNN outputs and demonstrate the validity of the proposed method using representative DNN-based medical image classifications. Black-box UAPs can be used to conduct both non-targeted and targeted attacks. Overall, the black-box UAPs showed high attack success rates (40% to 90%), although some of them had relatively low success rates because the method only utilizes limited information to generate UAPs. The vulnerability of black-box UAPs was observed in several model architectures. The results indicate that adversaries can also generate UAPs through a simple procedure under the black-box condition to foil or control DNN-based medical image diagnoses, and that UAPs are a more realistic security threat.

# Introduction

Adversarial examples [1–3] are input images that are typically generated by adding specific, imperceptible perturbations to the original input images, leading to a misclassification of deep neural networks (DNNs), thus questioning the generalization ability of a DNN, limiting its practical application in safety- and security-critical environments, and reducing the model interpretability [4–7]. In particular, because DNNs are powerful tools for image classification, and their diagnostic performance is high and equivalent to that of healthcare professionals, such networks are beginning to be applied to medical image diagnosis, empowering physicians and accelerating decision-making in clinical environments [8] [9]. However, because disease diagnosis involves making high-stake decisions, adversarial attacks can cause serious security problems [10] and various social problems [11]. The development of methods for generating adversarial perturbations is required to evaluate the reliability and safety of a DNN against adversarial attacks.

Many previous studies have considered input-dependent adversarial attacks (i.e., an individual adversarial perturbation is used such that each input image is misclassified). However, such adversarial attacks are difficult to achieve because they need to compute a different adversarial perturbation for each input (i.e., require high computational costs); thus, universal adversarial perturbations (UAPs) [12,13] and more realistic adversarial attacks. A UAP is a small single perturbation and can cause a failure of most image classification tasks of a DNN. Because UAPs are input-agnostic, adversaries can result in failed DNN-based classification tasks at lower costs using a UAP; specifically, they do not need to consider the distribution and diversity of the inputs. Moreover, UAPs are used not only for non-targeted attacks [12], which cause a misclassification (i.e., a task failure resulting in an input image being assigned an incorrect label), but also for targeted attacks [13], which cause the DNN to classify an input image into a specific class. UAP-based attacks can be more easily implemented by adversaries in real-world environments and are practicable [12–14].

However, widely used algorithms for generating UAPs [12,13] are limited to a white-box condition in which adversaries can access the model parameters (the gradient of the loss function in this case) and the training data. Given that DNN-based systems are generally operated under black-box conditions (i.e., as closed-source software and closed application programming interfaces in which only input queries are allowed and outputs are accessible) in terms of security, such universal adversarial attacks may be unrealistic. Several methods for black-box attacks that estimate the adversarial perturbations using only model outputs (e.g., confidence scores) have been proposed. For example, the zeroth-order optimization method [15] generates adversarial examples by estimating the loss gradients of a targeted DNN from their model outputs. Black-box adversarial perturbations are generatable in the context of optimization problems (i.e., minimizing the confidence score for the correct label for non-targeted attacks and maximizing the score for the target class for targeted attacks). As an example, a genetic algorithm-based method has been proposed [16]. Black-box attacks are executable with a low query cost by restricting the search for adversarial perturbations to a low-frequency domain [17]. Moreover, the simple black-box adversarial attack (SimBA) method considers a simple iterative search algorithm with random hill climbing to generate adversarial perturbation under the black-box condition [18]. However, these black-box attack methods are limited to input-dependent adversarial attacks. In addition, generative network models can be used to generate UAPs under the black-box condition [19]; however, they require a high computational cost (it is costly to train the network models). Black-box UAPs are also generatable with a relatively lower cost by considering that convolution neural networks (CNNs) are sensitive to the directions of the Fourier basis functions [20]. However, this method is limited to CNNs and may be less effective according to the CNN model architectures; moreover, it does not allow targeted attacks. A simpler and more effective algorithm for generating UAPs under black-box conditions is required to evaluate the reliability and safety of a DNN in more realistic environments.

Thus, herein, we propose a simple algorithm for generating black-box UPAs inspired by the SimBA method [18]. As mentioned above, this method considers a simple search to generate input-dependent adversarial perturbations under the black-box condition. Moreover, it can be used for both non-targeted and targeted attacks. We extended this earlier method to generate non-targeted and targeted UAPs. To demonstrate the validity of the proposed method, following our previous study [14], three representative medical image classifications were considered because adversarial attacks on medical machine learning are considered a security problem [11], i.e., skin cancer classification based on photographic images [21], referable diabetic retinopathy classification based on optical coherence tomography (OCT) images [22], and pneumonia classification based on chest X-ray images [22]. We obtained DNN models with several architectures for these medical image classifications and evaluated the vulnerability of the DNN models to black-box UAPs for both non-targeted and targeted attacks.

## Materials and Methods

### Black-box universal adversarial perturbations

Our algorithm (Algorithm 1) is an extension of the SimBA method [18] used to generate UAPs; in particular, we combined the SimBA method with simple iterative methods for generating UAPs under a white-box condition [12,13]. Similar to the SimBA method, our algorithm considers a black-box classifier $C$ that returns the confidence score output probability $p_C(y|x)$ of class $y$ given input image $x$. Here, $C(x)$ indicates the class or label (with the highest confidence score) for $x$ (i.e., $\arg\max_{y'} p_C(y'|x)$). The algorithm starts with $\boldsymbol{\delta} = \boldsymbol{0}$ (i.e., no perturbations) and iteratively updates the UAP $\boldsymbol{\delta}$ using a direction $\boldsymbol{q}$ randomly sampled from a set $\boldsymbol{Q}$ of search directions with attack strength $\epsilon$ under the constraint in which the $L_p$ norm of the UAP is equal to or less than a small value $\xi$ (i.e., $\|\boldsymbol{\delta}\|_p \leq \xi$). For project$(\boldsymbol{\delta}, \boldsymbol{p}, \xi)$, a projection function is used to satisfy the constraint, and is specifically defined as follows:

$$\text{project}(\boldsymbol{\delta}, p, \xi) = \arg\min_{\boldsymbol{\delta}^*} \|\boldsymbol{\delta} - \boldsymbol{\delta}^*\|_2 \quad \text{s.t.} \quad \|\boldsymbol{\delta}^*\|_p \leq \xi$$

For non-targeted attacks, the updated UAP vector $\boldsymbol{\delta}'$ is accepted if the update from $\boldsymbol{\delta}$ to $\boldsymbol{\delta}'$ contributes overall to decreasing the confidence scores for their labels $C(x)$ predicted without the UAP of any $x$ in an input image set $X$. As shown in Algorithm 1, the update

is accepted if the following simple condition is satisfied (true):

$$\sum_{x \in X} p_C(C(x)|x + \delta') < \sum_{x \in X} p_C(C(x)|x + \delta).$$

Targeted UAPs can also be implemented by modifying this condition (electronic supplementary material, Algorithm S1). Specifically, in this case, because the algorithm accepts an updated UAP vector if the update overall contributes to increasing the confidence score for a target class $y$ of any $x$ in $X$, the update is accepted if

$$\sum_{x \in X} p_C(y|x + \delta') > \sum_{x \in X} p_C(y|x + \delta).$$

Several types of search directions $Q$ were considered. As mentioned in a previous study [18], a natural choice for $Q$ may be the standard basis $Q = I$, for which $q$ corresponds to a random pixel attack. However, this may be less effective for inputs with a large space (i.e., large images). In this case, a discrete cosine basis is useful because random noise in a low-frequency space contributes to adversarial attacks [17]. The set $Q_{DCT}$ of the orthonormal frequencies was extracted using the discrete cosine transform. Assuming a two-dimensional image space $\mathbb{R}^{d \times d}$, $Q_{DCT}$ has $d \times d$ frequencies; however, a fraction $f_d$ of the lowest frequency directions is only used to generate the UAP more effectively (faster).

This update procedure terminates when the attack success rate for $X$ is 100% or the number of iterations reaches the maximum $i_{max}$. When generating a nontargeted UAP $\delta_{nt}$ the attack success rate corresponds to the fooling rate $R_f$, i.e., the fraction of adversarial images from which the labels predicted are inconsistent with the labels predicted from clean images to all images in set $X$, $R_f = |X|^{-1} \sum_{x \in X} \mathbb{I}(C(x) \neq C(x + \delta_{nt}))$, where the function $\mathbb{I}(A)$ returns 1 if condition $A$ is true and 0 otherwise. When generating a targeted UAP $\delta_t$, the attack success rate corresponds to the targeted attack success rate $R_s = |X|^{-1} \sum_{x \in X} \mathbb{I}(C(x + \delta_t) = y)$, i.e., the ratio of adversarial images classified into target class $y$ to all images in set $X$.

Our algorithm was implemented using the adversarial robustness toolbox (ART, version 1.7.0; github.com/Trusted-AI/adversarial-robustness-toolbox).

---

**Algorithm 1**: Computation of a nontargeted UAP

**Input:** Set $X$ of input images, classifier $C$, set $Q$ of search directions, attack strength $\epsilon$, cap $\xi$ on $L_p$ norm of the perturbation, norm type $p$ (1, 2, or $\infty$), maximum number $i_{max}$ of iterations.

**Output:** non-targeted UAP vector $\delta$

1:   $\delta \leftarrow 0, r \leftarrow 0, i \leftarrow 0$
2:   **while** $r < 1$ and $i < i_{max}$ **do**
3:       Pick a direction randomly: $q \in Q$
4:       **for** $\alpha \in \{-\epsilon, \epsilon\}$ **do**
5:          $\delta' \leftarrow \text{project}(\delta + \alpha q, p, \xi)$

```
 6:         if ∑_{x ∈ X} p_C(C(x)|x + δ') < ∑_{x ∈ X} p_C(C(x)|x + δ) then
 7:             δ ← δ'
 8:             break
 9:         end if
10:     end for
11:     r ← |X|^{-1} ∑_{x∈X} 𝕀(C(x) ≠ C(x + δ))
12:     i ← i + 1
13: end while
```

*Medical images and DNN models*

To evaluate the validity of the proposed method, DNN-based medical image classification is considered. We used the medical image datasets and DNN models previously described in [14] (see also github.com/hkthirano/MedicalAI-UAP). A brief description is provided below.

Skin lesion images for skin cancer classification are classified into seven classes: melanoma (MEL), melanocytic nevus (NV), basal cell carcinoma (BCC), actinic keratosis/Bowens disease (intraepithelial carcinoma, AKIEC), benign keratosis (solar lentigo/seborrheic keratosis/lichen planus-like keratosis, BKL), dermatofibroma (DF), and vascular lesions (VASC). The OCT images for referable diabetic retinopathy classification are classified into four classes: choroidal neovascularization with neovascular membrane and associated subretinal fluid (CNV), diabetic macular edema with retinal-thickening-associated intraretinal fluid (DME), multiple drusen present in early age-related macular degeneration (DRUSEN), and normal retina with preserved foveal contour and absence of any retinal fluid/edema (NM). The chest X-ray images for pneumonia classification are classified into binary classes: no pneumonia (NORMAL) or viral or bacterial pneumonia (PNEUMONIA). All images had a pixel resolution of 299 × 299. The skin lesion images are red–green–blue, whereas the OCT and chest X-ray images are in grayscale. We used the test images to generate UAPs and evaluate the UAP performance because the black-box condition assumes that adversaries cannot access the training data. In the skin lesion, OCT, and chest X-ray image datasets, 3,015, 3,360, and 540 test images were available, respectively. Note that the OCT and chest X-ray image datasets were class-balanced, whereas the skin lesion image dataset was not.

To evaluate the effect of the model architecture on vulnerability to UAPs, we used the Inception V3 architecture [23], VGG16 [24], and ResNet50 [25] architectures. The DNN models were obtained using transfer learning, and the DNN model architectures pre-trained using the ImageNet dataset [26] were fine-tuned with the training images in a medical image dataset, using the learning rate schedule and data augmentation (see [14] for the test accuracies of the DNN models).

*Generating UAPs*

Non-targeted and targeted UAPs were generated using a portion of the test images in the dataset. For the OCT image dataset, 800 test images (200 randomly selected images per class) were used. For the chest X-ray image dataset, 200 test images (100 images randomly selected per class) were used. For the skin lesion image dataset, 1,000 randomly selected test images were used. The parameters $\epsilon$ and $i_{\max}$ were set to 0.5 and 5,000, respectively. As the set of search directions, $\boldsymbol{Q}_{\mathrm{DCT}}$ was considered, where $f_d$ was set to ~9.4% (= 28/299). The parameter $\xi$ was set based on the ratio $\zeta$ of the $L_p$ norm of the UAP to the average $L_p$ norm of an image in the dataset (see [14] for the actual values of the average $L_p$ norms). Following [14], in addition, $\zeta = 4\%$ for the skin lesion and chest X-ray image dataset, and $\zeta = 8\%$ for the OCT image dataset.

*Evaluating the performance of UAPs*

The performances of a nontargeted UAP and a targeted UAP were evaluated using the fooling rate $R_f$ and targeted attack success rate $R_s$, respectively. In addition, $R_f$ and $R_s$ of a UAP were computed using the validation dataset, which consists of the rest of the test images (i.e., the test images excluded the images used to generate the UAP) in each medical image dataset. However, when evaluating the effect of the number of images used to generate a UAP on the performance of the UAP, a validation dataset of the same size was used to evaluate the performance, and a fixed number of test images were randomly selected from the remaining images (validation dataset) for each medical dataset. To compare the performances of the generated UAPs with those of the random controls, random vectors (random UAPs) were sampled uniformly from the sphere of a specified radius [12]. As mentioned in our previous study [14], note that $R_s$ has a baseline and is observed without UAPs. The class (label) composition of the image data and prediction performance of DNNs both affect the baseline. In this study, the $R_s$ baselines of UAPs targeted to a specified class were ~25% and ~50%, respectively, for the OCT and chest X-ray image datasets. For the skin lesion dataset, the $R_s$ baselines of UAPs targeted to MEL and NV were ~10% and ~65%, respectively. In addition, $R_f$ and $R_s$ were computed using test images from a medical image dataset.

The confusion matrices on the validation dataset (i.e., the rest of the test images used to generate a UAP) were also obtained to evaluate the change in prediction owing to the UAPs for each class. The rows and columns in the matrices represent the true and predicted classes, respectively. The confusion matrices were row-normalized to account for imbalanced datasets.

## Results

*Non-targeted universal adversarial attacks*

We first evaluated the performance of UAP-based non-targeted attacks on DNN-based medical image classifications (Table 1). Overall, the fooling rate $R_f$ of the UAPs was significantly higher than that of the random UAPs (random controls). Moreover, the UAPs were almost imperceptible. As a representative example, Figure 1 shows both clean images and their adversarial versions through the use of UAPs against the Inception V3 models because this model architecture has been widely used in previous studies on DNN-based medical imaging (e.g., [21,22]), where $p = 2$ for the skin lesion and chest X-ray image datasets, and $p = \infty$ for the OCT image dataset in terms of the UAP performance. The results indicate that small UAPs are generatable under black-box conditions. However, the performance ($R_f$) might depend on the model architectures and norm type $p$ of the UAPs for each medical image dataset. For the skin lesion image dataset, $R_f$ achieved a score of >70% when $p = 2$ regardless of the model architecture; however, it was relatively low (~35%) for the UAPs with $p = \infty$ against the ResNet50 and VGG16 models. A similar $R_f$ (~65%) was also observed for the Inception V3 model when $p = \infty$. For the OCT image dataset, $R_f$ of the UAP with $p = \infty$ was higher than that of the UAP with $p = 2$; in particular, the UAP with $p = 2$ against the ResNet50 model was less effective in causing a misclassification of the DNN model, although it still showed a slightly higher $R_f$ in comparison to random UAPs. For the chest X-ray image dataset, $R_f$ (~40%) of the UAPs against the Inception V3 model was slightly lower than that (~50%) of the UAPs against the ResNet50 and VGG16 models, regardless of the norm type.

The confusion matrices (Fig. 2) show that the models classified many images into several specific classes (i.e., dominant classes) owing to the UAPs; however, the dominant classes might differ according to the model architecture. For the skin lesion image dataset, the dominant class was BKL for the Inception V3 model, whereas it was MEL for the ResNet50 model. For the VGG model, many images were classified as BCC and BKL. For the OCT image classification, the dominant class was CNV for the ResNet50 and VGG16 models; by contrast, many images were misclassified as either CNV or DME for the Inception V3 model. For the chest X-ray image dataset, regardless of the model architecture, most images were classified as PNEUMONIA because of the UAPs, indicating that $R_f$ saturated at ~50% (Table 1).

*Targeted universal adversarial attacks*

Next, we evaluated the performance of UAP-based targeted attacks on the DNN-based medical image classifications (Table 2), where $p = 2$ for the skin lesion and chest X-ray image datasets, and $p = \infty$ for the OCT image dataset, when considering the performance of non-targeted UAPs (Table 1). Following our previous study [14], we considered the targeted attacks to be the most significant cases, which were used as a control in each medical image dataset. The most significant cases correspond to MEL, CNV, and PNEUMONIA in the skin lesion, OCT, and chest X-ray image datasets, respectively. In addition, the controls correspond to NV, NM, and NORMAL in the skin lesion, OCT, and chest X-ray image datasets, respectively. Table 1 shows the targeted attack success rate $R_s$ of the UAPs against DNN-based medical image classification. The UAPs were almost imperceptible because $\zeta = 4\%$ in the skin lesion and chest X-ray image datasets, and $\zeta = 8\%$ in the OCT image dataset, as in the case of non-targeted UAPs (see Fig. 1).

Overall, the targeted attacks on the most significant cases were successful for all types of medical image classification. The values of $R_s$ (60% to 95%) of the UAPs were significantly higher than those of the random controls. Here, $R_s$ was independent of the model architecture for the OCT and chest X-ray image datasets. However, a weak model architecture dependency of $R_s$ was observed for the skin lesion image dataset. The value of $R_s$ (~60%) of the UAP against the Inception V3 model was slightly lower than that (~80%) of the UAP against the ResNet50 and VGG16 models.

By contrast, attacks targeted against the controls were only partly successful. For the skin lesion image dataset, $R_s$ of the UAP for the attacks targeted against NV were 75% to 80% for the ResNet50 and VGG16 models, which is significantly higher than that of the random controls; however, it is similar to that of the random control for Inception V3, indicating that the targeted attacks failed. For the OCT image dataset, the values of $R_s$ of the UAP for attacks targeted against NM were also almost equivalent to those of random UAPs regardless of the model architecture. For the chest X-ray image dataset, the performance of UAP-based attacks targeted against NORMAL was remarkably high for the VGG16 model ($R_s$ ~95%); however, $R_s$ was relatively low (~65%) for the Inception V3 model compared to the UAP against the VGG16 model, although $R_s$ was higher than that of the random control. For the ResNet50 model, $R_s$ was similar to the random control.

*Effect of input dataset size on the UAP performance*

Finally, we investigated the effect of the size of the input dataset used to generate a UAP on the UAP performance. As a representative example, in terms of such performance, we focused on the skin lesion image dataset and evaluated the performance of nontargeted UAPs with $p = 2$. The UAPs were generated using an input dataset consisting of $N$ images randomly selected from the test images in the medical image dataset. Here, we considered $N = 100$, $N = 500$, and $N = 1000$. The value of $R_f$ was computed using the input and validation datasets. The validation dataset consisted of 2015 images randomly selected from the test images, excluding the images in the input dataset.

As shown in Fig. 3, $R_f$ increased with $N$. In addition, $R_f$ for the validation dataset was lower than that for the input dataset; however, the difference in $R_f$ between the validation dataset and input dataset decreased with an increase in $N$. These tendencies were observed regardless of the model architecture. The results indicate that UAPs achieve a high performance and generalization, resulting in a misclassification of the DNN model with an increasing $N$. However, the UAPs showed a relatively high performance despite a small $N$. When $N = 100$, $R_f$ for the validation dataset was > 50% for the Inception V3 and VGG models, although it was ~30% for the ResNet50 model. Moreover, $R_f$ for the validation dataset was >60% regardless of the model architecture when $N = 500$.

## Discussion

We proposed a simple method for generating UAPs under the black-box condition inspired by the SimBA method [18] and applied this method to DNN-based medical image classification. Overall, the results showed that small (almost imperceptible) UAPs are generatable under the black-box condition only using a simple hill climbing search based on the model outputs (i.e., confidence scores) to perform both non-targeted and targeted attacks. The vulnerability of black-box UAPs was observed in several model architectures, indicating the versatility of our method; thus, it may be a general property of a DNN. The results indicate that adversaries can foil or control DNN-based medical image diagnoses even if they never access the model parameters (e.g., loss gradients) and training data because target systems are operated under the black-box condition in terms of security. DNN-based medical image diagnoses may be easy to conduct in a realistic environment.

It can be argued that the performance of black-box UAPs generated by our method compares favorably with that of white-box UAPs [12,13], given that, in addition to the model parameters, the white-box UAPs were generated using a larger dataset (7–10 times) than that of the black-box UAPs. For instance, for the non-targeted UAPs with $p = 2$ and $\zeta = 4\%$ against the Inception V3 model for the skin lesion image dataset, $R_f$ of the black-box UAP was ~80% (Table 1), whereas that of the white-box UAP was ~90% (see Table 1 in [14]), where 1000 test images and 7000 training images were used to generate the black-box and white-box UAPs, respectively. For non-targeted attacks, overall, the values of $R_f$ of the black-box UAPs are 40% to 80% (Table 1, except for the nonsignificant cases in which $R_f$ is almost similar to the random control), whereas $R_f$ of the white-box UAPs is 70% to 90% [14]. For targeted attacks, overall, the values of $R_s$ of the black-box UAPs are 60% to 90% (Table 2, except for the nonsignificant cases), whereas those of the white-box UAPs are >90% (see Table 2 in [14]).

Our proposed method generated UAPs with a relatively high performance despite a small number of queries because it is based on the SimBA method [18], which is a query efficient approach. However, in addition to the advantages mentioned in the Introduction, black-box UAPs have an advantage in that adversaries do not need to be concerned with the query efficiency in black-box attack methods. In general, the number of queries per fixed time is limited in terms of software and application programming interfaces in terms of security; thus, query-efficient methods are needed for input-dependent adversarial attacks. However, the number of queries required for generating UAPs poses few problems, even if they are large because UAPs are input-agnostic.

Black-box UAPs are likely useful for avoiding adversarial defenses. For example, a discontinuous activation function is often used to make it difficult for adversaries to estimate the loss gradients [27]; however, black-box attack methods have led to many defense methods [28], including the use of a discontinuous activation function. Black-box attacks based on UAPs may provide insight into the development of more efficient defense methods.

However, the performance of black-box UAPs is limited in some cases. For non-targeted attacks on the models for the chest X-ray image dataset, the values $R_f$ of the black-box UAPs were at most ~50% (most images were classified into PNEUMONIA only; Table 1 and Fig. 2), whereas those of the white-box UAPs were at most ~80% (the models incorrectly predicted the true labels; Table 1 in [14]). In addition, the success rates of the targeted attacks to the controls (i.e., NV, NM, and NORMAL for the skin lesion, OCT, and chest X-ray image datasets, respectively) were mostly equivalent to the random controls (Table 2), although the white-box UAPs achieved a high performance (Table 2 in [14]). This is because the black-box attack method utilizes only limited information to generate UAPs, compared to white-box methods. Moreover, this may be due to an imbalanced dataset. NV images were abundant in the skin lesion image dataset. Because the proposed algorithm considers maximizing the confidence score for a targeted class, a large $R_s$ will have been already been achieved for targeted attacks to an abundant label in a dataset; thus, UAPs are rarely updated in the algorithm, and as a result, $R_s$ rarely increases. Simple solutions for improving the performance include the use of more input images (e.g., as shown in Fig. 3 and using data augmentation) and considering a larger number of queries.

Another solution is to consider different types of search directions. For example, the Fourier basis [20], texture bias [29] and Turing patterns [30] may be useful for efficiently searching for UAPs because they are also useful for universal adversarial attacks.

Our proposed method is limited to a black-box condition in which the confidence scores for all labels are available. A harder black-box condition can also be considered, i.e., a case in which the classifiers return the predicted label only. For example, the boundary attack method [31] generates input-dependent adversarial perturbations based on a decision-based attack that starts from a large adversarial perturbation and then seeks to reduce the perturbations while remaining adversarial. This method requires a relatively large number of model queries; however, the HopSkipJumpAttack method [32] can conduct decision-based adversarial attacks with significantly fewer model queries by using binary information at the decision boundary. Nevertheless, this study considered confidence-score-based black-box attacks because the use of confidence scores is important in deciding whether to trust a classifier's decision in terms of machine learning trust and safety [33] (for healthcare in particular [34,35]). However, it is important to evaluate whether adversarial attacks are possible under hard black-box conditions in terms of the reliability and safety of a DNN. Although further investigations are needed, our algorithm may provide insight into the development of decision-based universal adversarial attacks.

In summary, we proposed a simple method for generating UAPs under black-box conditions and demonstrated that black-box UAPs can be generated easily using a relatively small dataset. Our study enhances our understanding of the vulnerabilities of a DNN to adversarial attacks. Moreover, our finding that adversaries can generate UAPs under the black-box condition using a simple procedure provides insight into increasing the reliability and safety of a DNN and designing its operational strategy (for medical imaging in particular [10]).

# References

1. Goodfellow, I. J., Shlens, J. & Szegedy, C. 2015 Explaining and harnessing adversarial examples. *Int. Conf. Learn. Represent.*

2. Yuan, X., He, P., Zhu, Q. & Li, X. 2019 Adversarial examples: attacks and defenses for deep learning. *IEEE Trans. Neural Networks Learn. Syst.* **30**, 2805–2824. (doi:10.1109/TNNLS.2018.2886017)

3. Ortiz-Jimenez, G., Modas, A., Moosavi-Dezfooli, S.-M. & Frossard, P. 2020 Optimism in the face of adversity: Understanding and improving deep learning through adversarial robustness.

4. Matyasko, A. & Chau, L.-P. 2018 Improved network robustness with adversary critic. *Proc. 32nd Int. Conf. Neural Inf. Process. Syst.* , 10601–10610.

5. Ortiz-Jimenez, G., Modas, A., Moosavi-Dezfooli, S.-M. & Frossard, P. 2021 Optimism in the Face of Adversity: Understanding and Improving Deep Learning Through Adversarial Robustness. *Proc. IEEE* **109**, 635–659. (doi:10.1109/JPROC.2021.3050042)

6. Madry, A., Makelov, A., Schmidt, L., Tsipras, D. & Vladu, A. 2018 Towards deep learning models resistant to adversarial attacks. In *International Conference on Learning Representations*,

7. Carlini, N. & Wagner, D. 2017 Towards evaluating the robustness of neural networks. *Proc. - IEEE Symp. Secur. Priv.* , 39–57. (doi:10.1109/SP.2017.49)

8. Litjens, G., Kooi, T., Bejnordi, B. E., Setio, A. A. A., Ciompi, F., Ghafoorian, M., van der Laak, J. A. W. M., van Ginneken, B. & Sánchez, C. I. 2017 A survey on deep learning in medical image analysis. *Med. Image Anal.* **42**, 60–88. (doi:10.1016/j.media.2017.07.005)

9. Liu, X. et al. 2019 A comparison of deep learning performance against health-care professionals in detecting diseases from medical imaging: a systematic review and meta-analysis. *Lancet Digit. Heal.* **1**, e271–e297. (doi:10.1016/S2589-7500(19)30123-2)

10. Kaissis, G. A., Makowski, M. R., Rückert, D. & Braren, R. F. 2020 Secure, privacy-preserving and federated machine learning in medical imaging. *Nat. Mach. Intell.* **2**, 305–311. (doi:10.1038/s42256-020-0186-1)

11. Finlayson, S. G., Bowers, J. D., Ito, J., Zittrain, J. L., Beam, A. L. & Kohane, I. S. 2019 Adversarial attacks on medical machine learning. *Science (80-. ).* **363**, 1287–1289. (doi:10.1126/science.aaw4399)

12. Moosavi-Dezfooli, S. M., Fawzi, A., Fawzi, O. & Frossard, P. 2017 Universal adversarial perturbations. *Proc. - 30th IEEE Conf. Comput. Vis. Pattern*

*Recognition, CVPR 2017* **2017-Janua**, 86–94. (doi:10.1109/CVPR.2017.17)

13. Hirano, H. & Takemoto, K. 2020 Simple iterative method for generating targeted universal adversarial perturbations. *Algorithms* **13**, 268. (doi:10.3390/a13110268)

14. Hirano, H., Minagi, A. & Takemoto, K. 2021 Universal adversarial attacks on deep neural networks for medical image classification. *BMC Med. Imaging* **21**, 9. (doi:10.1186/s12880-020-00530-y)

15. Chen, P.-Y., Zhang, H., Sharma, Y., Yi, J. & Hsieh, C.-J. 2017 ZOO: Zeroth Order Optimization based black-box attacks to deep neural networks without training substitute models. In *Proceedings of the 10th ACM Workshop on Artificial Intelligence and Security*, pp. 15–26. New York, NY, USA: ACM. (doi:10.1145/3128572.3140448)

16. Chen, J., Su, M., Shen, S., Xiong, H. & Zheng, H. 2019 POBA-GA: Perturbation optimized black-box adversarial attacks via genetic algorithm. *Comput. Secur.* **85**, 89–106. (doi:10.1016/j.cose.2019.04.014)

17. Guo, C., Frank, J. S. & Weinberger, K. Q. 2019 Low frequency adversarial perturbation. In *Proceedings of the Thirty-Fifth Conference on Uncertainty in Artificial Intelligence, {UAI} 2019, Tel Aviv, Israel, July 22-25, 2019* (eds A. Globerson & R. Silva), pp. 1127–1137. {AUAI} Press.

18. Guo, C., Gardner, J. R., You, Y., Wilson, A. G. & Weinberger, K. Q. 2019 Simple black-box adversarial attacks. *Proc. 36th Int. Conf. Mach. Learn.* , 2484–2493.

19. Poursaeed, O., Katsman, I., Gao, B. & Belongie, S. 2018 Generative Adversarial Perturbations. In *2018 IEEE/CVF Conference on Computer Vision and Pattern Recognition*, pp. 4422–4431. IEEE. (doi:10.1109/CVPR.2018.00465)

20. Tsuzuku, Y. & Sato, I. 2019 On the structural sensitivity of deep convolutional networks to the directions of Fourier basis functions. In *IEEE Conference on Computer Vision and Pattern Recognition, CVPR 2019, Long Beach, CA, USA, June 16-20, 2019*, pp. 51–60. Computer Vision Foundation / {IEEE}. (doi:10.1109/CVPR.2019.00014)

21. Esteva, A., Kuprel, B., Novoa, R. A., Ko, J., Swetter, S. M., Blau, H. M. & Thrun, S. 2017 Dermatologist-level classification of skin cancer with deep neural networks. *Nature* **542**, 115–118. (doi:10.1038/nature21056)

22. Kermany, D. S. et al. 2018 Identifying medical diagnoses and treatable diseases by image-based deep learning. *Cell* **172**, 1122-1131.e9. (doi:10.1016/j.cell.2018.02.010)

23. Szegedy, C., Vanhoucke, V., Ioffe, S., Shlens, J. & Wojna, Z. 2016 Rethinking the Inception architecture for computer vision. In *2016 IEEE Conference on Computer Vision and Pattern Recognition (CVPR)*, pp. 2818–2826. IEEE. (doi:10.1109/CVPR.2016.308)

24. Simonyan, K. & Zisserman, A. 2015 Very deep convolutional networks for large-scale image recognition. In *3rd International Conference on Learning Representations, ICLR 2015 - Conference Track Proceedings*,

25. He, K., Zhang, X., Ren, S. & Sun, J. 2016 Deep residual learning for image recognition. In *2016 IEEE Conference on Computer Vision and Pattern Recognition (CVPR)*, pp. 770–778. IEEE. (doi:10.1109/CVPR.2016.90)

26. Russakovsky, O. et al. 2015 ImageNet large scale visual recognition challenge. *Int. J. Comput. Vis.* (doi:10.1007/s11263-015-0816-y)

27. Xiao, C., Zhong, P. & Zheng, C. 2020 Enhancing adversarial defense by k-winners-take-all. *Proc. 8th Int. Conf. Learn. Represent.*

28. Mahmood, K., Gurevin, D., van Dijk, M. & Nguyen, P. H. 2020 Beware the black-box: On the robustness of recent defenses to adversarial examples.

29. Geirhos, R., Rubisch, P., Michaelis, C., Bethge, M., Wichmann, F. A. & Brendel, W. 2019 ImageNet-trained CNNs are biased towards texture; increasing shape bias improves accuracy and robustness. In *International Conference on Learning Representations*,

30. Tursynbek, N., Vilkoviskiy, I., Sindeeva, M. & Oseledets, I. V 2021 Adversarial Turing patterns from cellular automata. In *Thirty-Fifth AAAI Conference on Artificial Intelligence*, pp. 2683–2691. {AAAI} Press.

31. Brendel, W., Rauber, J. & Bethge, M. 2018 Decision-based adversarial attacks: Reliable attacks against black-Box machine learning models. In *International Conference on Learning Representations*,

32. Chen, J., Jordan, M. I. & Wainwright, M. J. 2020 HopSkipJumpAttack: A query-efficient decision-based attack. In *2020 IEEE Symposium on Security and Privacy (SP)*, pp. 1277–1294. IEEE. (doi:10.1109/SP40000.2020.00045)

33. Jiang, H., Kim, B., Guan, M. & Gupta, M. 2018 To trust or not to trust a classifier. In *Advances in Neural Information Processing Systems* (eds S. Bengio H. Wallach H. Larochelle K. Grauman N. Cesa-Bianchi & R. Garnett), Curran Associates, Inc.

34. Amann, J., Blasimme, A., Vayena, E., Frey, D. & Madai, V. I. 2020 Explainability for artificial intelligence in healthcare: a multidisciplinary perspective. *BMC Med. Inform. Decis. Mak.* **20**, 310. (doi:10.1186/s12911-020-01332-6)

35. Lauritsen, S. M., Kristensen, M., Olsen, M. V., Larsen, M. S., Lauritsen, K. M., Jørgensen, M. J., Lange, J. & Thiesson, B. 2020 Explainable artificial intelligence model to predict acute critical illness from electronic health records. *Nat. Commun.* **11**, 3852. (doi:10.1038/s41467-020-17431-x)

## Acknowledgments

We would like to thank Editage (www.editage.jp) for the English language editing.

## Ethical Statement

This study required no ethical permit.

## Funding Statement

This research was funded by JSPS KAKENHI (grant number 21H03545).

## Data Accessibility

All data generated and analyzed during this study are included in this published article and its supplementary information files. The code and data used in this study are available from the GitHub repository at github.com/kztakemoto/U-SimBA.

## Competing interests

The authors declare no competing interests.

## Authors' contributions

KK and KT conceived and designed the study. KK prepared the data and model. KK and KT coded and conducted the experimental evaluations. KK and KT interpreted the results. KK and KT wrote the manuscript. All authors have approved the final manuscript for publication.

# Figure captions

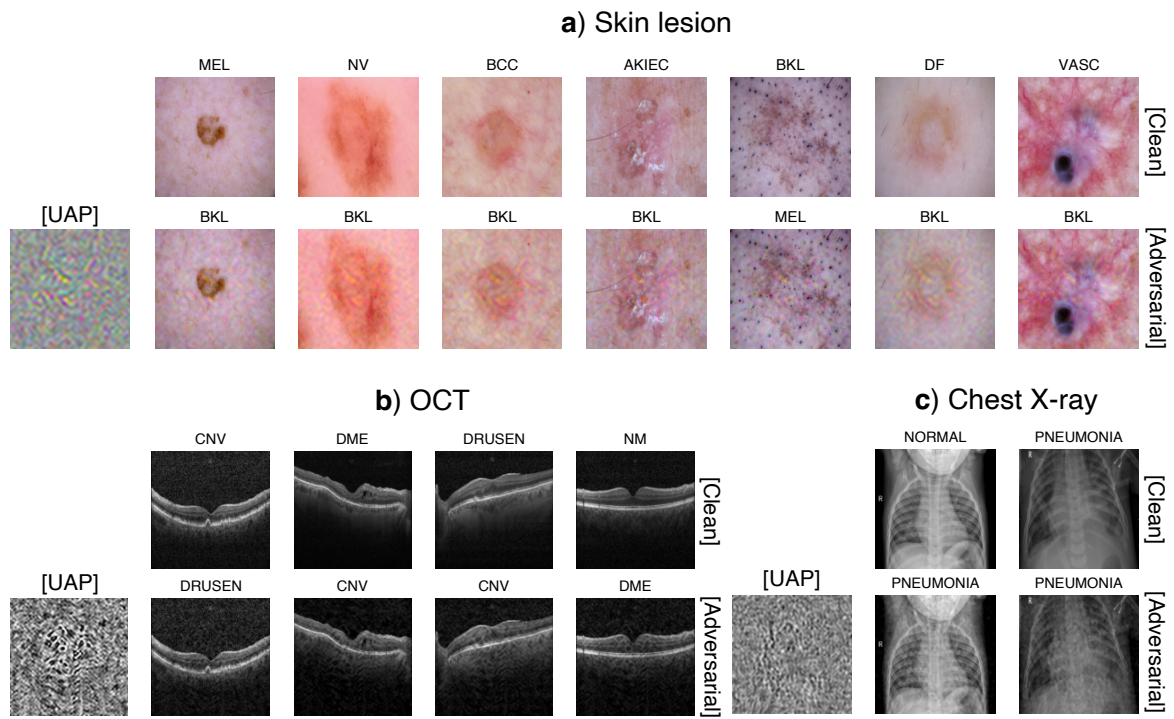

**Figure 1:** Clean images and their adversarial examples generated using nontargeted UAPs from the ImageNet dataset, the against Inception V3 model for the (a) skin lesion, (b) OCT, and (c) chest X-ray image classifications. Here, $p = 2$ in (a) and (c) $p = \infty$ in (b). Labels (without square brackets) next to the images are the predicted classes. The clean (original) images are correctly classified into their actual labels. UAPs are visually emphasized for clarity; in particular, each UAP is scaled by a maximum of 1 and a minimum of zero.

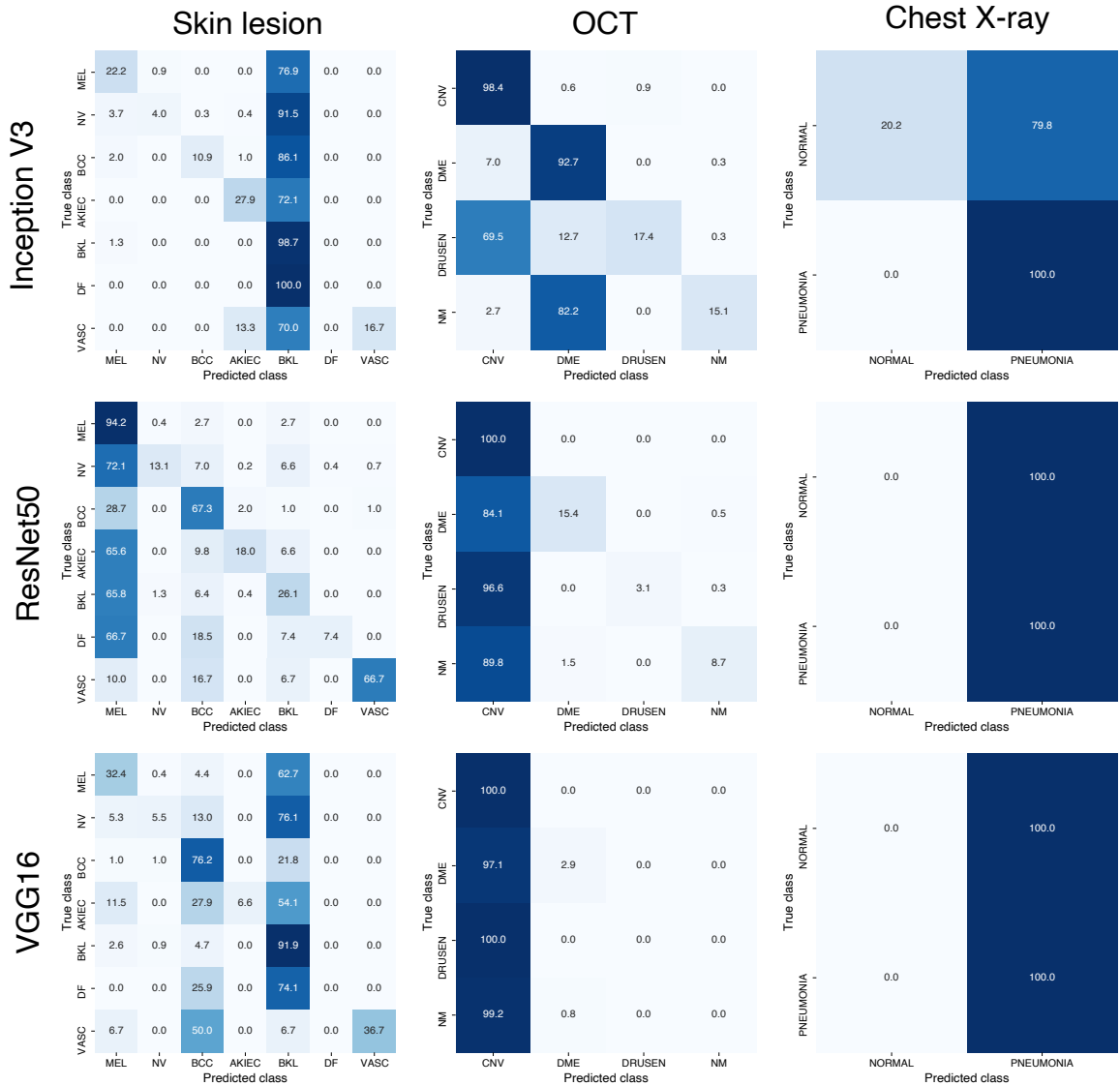

**Figure 2:** Normalized confusion matrices for Inception V3, ResNet50, and VGG16 models attacked using nontargeted UAPs for skin lesions, OCT, and chest X-ray image datasets. Here, $p = 2$ for skin lesion and chest X-ray image datasets, and $p = \infty$ for the OCT image dataset.

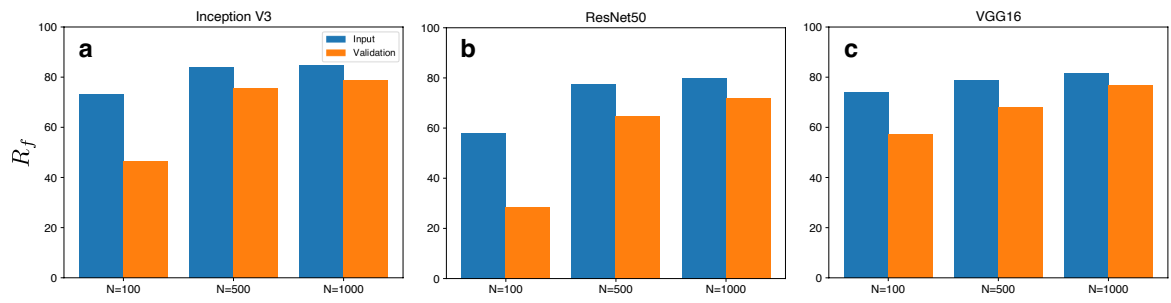

**Figure 3:** Effect of input dataset size on the fooling rates $R_f$ (%) of UAPs against Inception V3 (a), ResNet50 (b), VGG16 models (c) for skin lesion image dataset. The values of $R_f$ for both the input and validation datasets are shown.

# Tables

**Table 1:** Fooling rates $R_f$ (%) of nontargeted UAPs against Inception V3, ResNet50, and VGG models for skin lesions, OCT, and chest X-ray image datasets. Values in brackets are $R_f$ of random UAPs (random controls).

| Model architecture | Skin lesion | | OCT | | Chest X-ray | |
|---|---|---|---|---|---|---|
| | $p = 2$ | $p = \infty$ | $p = 2$ | $p = \infty$ | $p = 2$ | $p = \infty$ |
| Inception V3 | 78.8 (13.6) | 65.6 (10.2) | 31.7 (1.6) | 44.9 (3.3) | 41.8 (2.1) | 44.1 (2.6) |
| ResNet50 | 71.9 (11.1) | 33.9 (8.6) | 5.5 (1.3) | 69.3 (4.3) | 51.5 (5.9) | 50.9 (6.2) |
| VGG16 | 76.6 (5.3) | 38.9 (3.6) | 40.9 (0.7) | 75.1 (2.0) | 50.0 (1.8) | 50.0 (2.4) |

**Table 2:** Target attack success rates $R_s$ (%) of targeted UAPs against Inception V3, ResNet50, and VGG models for skin lesions, OCT, and chest X-ray image datasets. $p = 2$ for skin lesion and chest X-ray datasets, whereas $p = \infty$ for OCT image dataset. Values in brackets are $R_s$ of random UAPs (random controls).

| Model architecture / target class | Skin lesion | | OCT | | Chest X-ray | |
|---|---|---|---|---|---|---|
| | NV | MEL | NM | CNV | NORMAL | PNEUMONIA |
| Inception V3 | 63.8 (64.8) | 60.9 (10.4) | 27.3 (27.2) | 92.2 (25.2) | 67.1 (54.4) | 91.8 (45.9) |
| ResNet50 | 76.0 (66.6) | 81.2 (10.6) | 28.6 (28.3) | 93.5 (24.8) | 53.8 (57.6) | 97.6 (42.4) |
| VGG16 | 80.0 (72.4) | 79.0 (7.7) | 25.1 (26.5) | 97.9 (24.6) | 97.4 (51.5) | 97.1 (48.5) |